%% file: main.tex
\documentclass[10pt,twocolumn,letterpaper]{article}

\usepackage{booktabs}   
\usepackage{makecell}   
\usepackage{multirow}   
\usepackage{graphicx}   
\usepackage{subfigure}
\usepackage{booktabs}
\usepackage{multirow} 
\usepackage{cvpr}              
\input{preamble}
\definecolor{cvprblue}{rgb}{0.21,0.49,0.74}
\usepackage[pagebackref,breaklinks,colorlinks,allcolors=cvprblue]{hyperref}

\def\paperID{10} 
\def\confName{CVPR}
\def\confYear{2026}

\title{SPR-128K: A New Benchmark for Spatial Plausibility Reasoning with Multimodal Large Language Models}

\author{Zhiyuan Hu\\
Tsinghua University\\
No. 30 Shuangqing Road, Haidian District, Beijing, China, 100084\\
{\tt\small huzy24@mails.tsinghua.edu.cn}
\and
Zheng Sun, Yi Wei, Long Yu\\
Alibaba Health Information Technology Limited\\
Building 9, Block 4, Wangjing East Park, Chaoyang District, Beijing, China, 100102\\
{\tt\small banqun.sz@alibaba-inc.com, wy271630@alibaba-inc.com, yl185725@alibaba-inc.com}
}

\begin{document}
\maketitle
\input{sec/0_abstract}    
\input{sec/1_intro}

{
    \small
    \bibliographystyle{ieeenat_fullname}
    \bibliography{main}
}


\end{document}

%% file: sec/0_abstract.tex
\begin{abstract}
The performance of image generation has been significantly improved in recent years. However, the study of image screening is rare, and its performance with Multimodal Large Language Models (MLLMs) is unsatisfactory due to the lack of data and the weak spatial plausibility reasoning ability in MLLMs. In this work, we propose a complete solution to address these problems in terms of data and methodology. For data, we collect a comprehensive spatial plausibility reasoning (SPR) dataset with over 128k samples, called SPR-128K. The dataset evaluates spatial plausibility reasoning ability under four aspects. Regarding data annotation, we investigate multiple approaches to acquire high-quality Chain-of-Thought (CoT) data in the most cost-effective manner. Methodologically, we introduce a Dynamic Proportional Accuracy (DPA) reward into the Group Relative Policy Optimization (GRPO) framework, called DPA-GRPO. This enhanced method demonstrates superior performance compared to the original GRPO. Our experiments reveal that even leading MLLMs exhibit unsatisfactory performance in spatial plausibility reasoning. In contrast, our much smaller model, leveraging DPA-GRPO, substantially surpasses both large open-source and leading closed-source models.
\end{abstract}

%% file: sec/1_intro.tex
\section{Introduction}
\label{sec:intro}

\begin{figure}[!t]
\centering

\subfigure[Four evaluation dimensions of spatial plausibility.]{
    \includegraphics[width=0.90\linewidth]{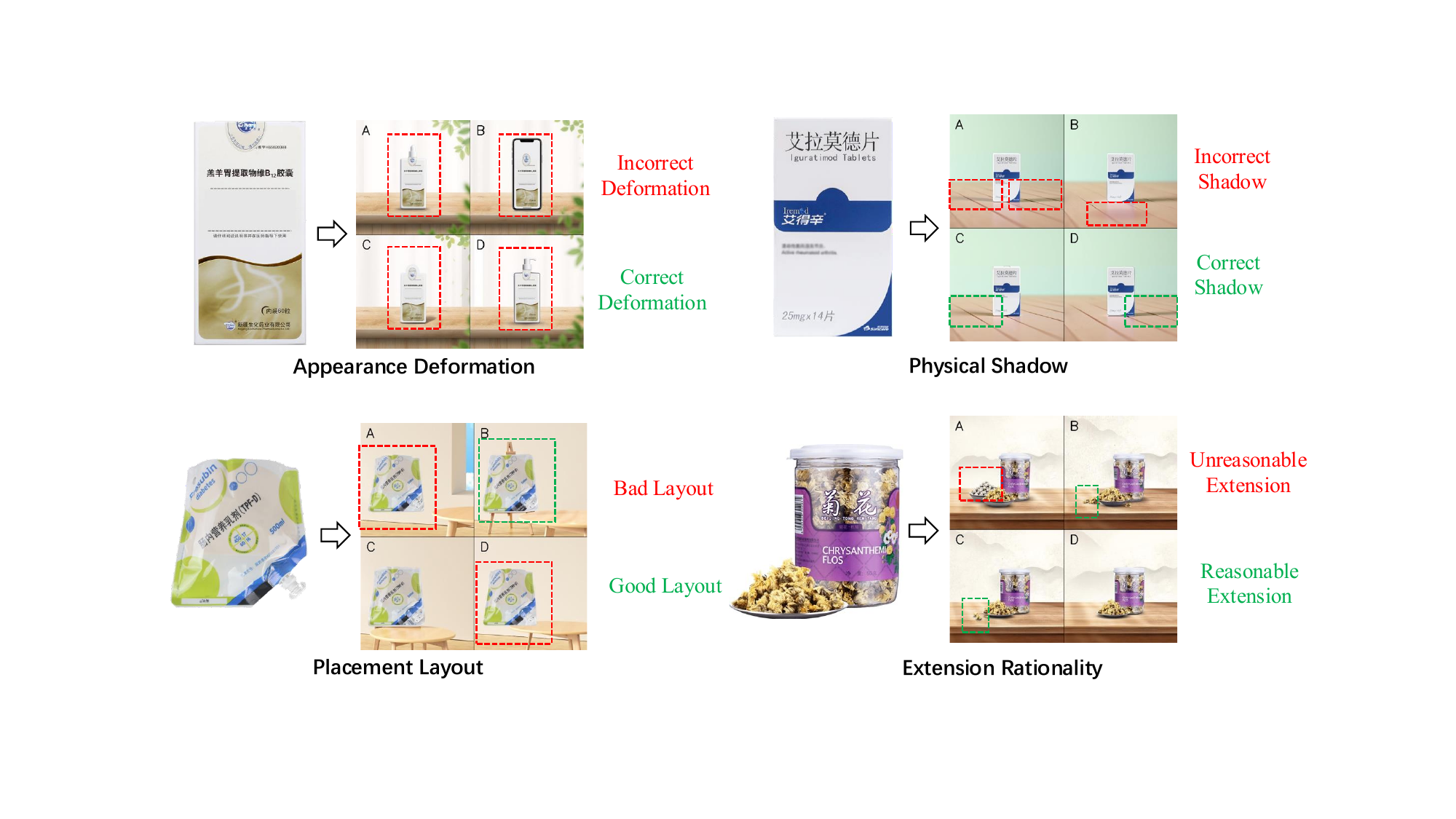}
}%
\quad
\subfigure[Quantitative comparison results.]{
    \includegraphics[width=0.90\linewidth]{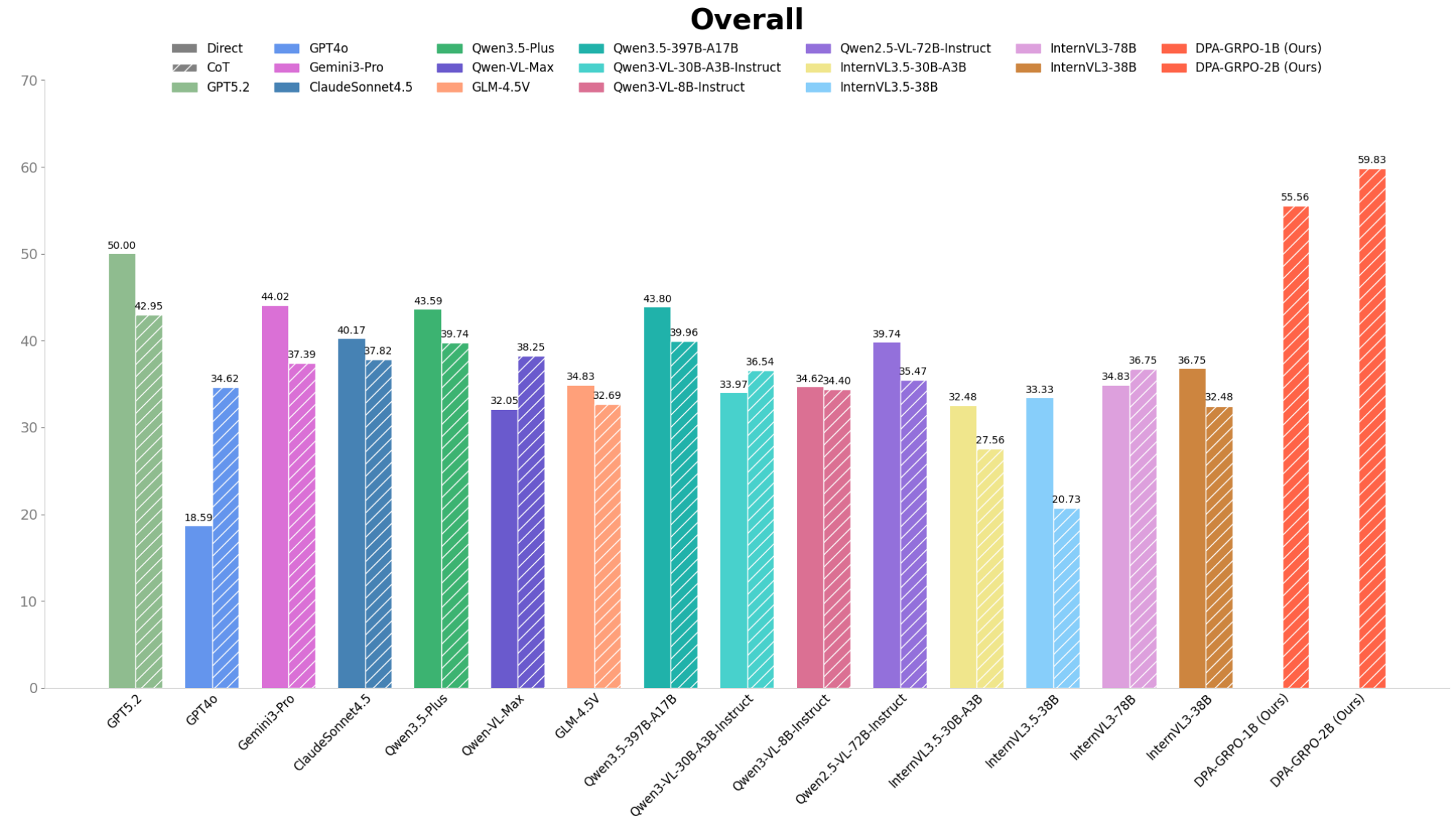}
}%

\centering
\caption{Overview of the spatial plausibility dataset and quantitative comparison results. (a) We summarize four evaluation dimensions of spatial plausibility from the dataset. (b) Extensive quantitative comparison results demonstrate the superiority of our DPA-GRPO method in the image screening task.} \label{Overview}
\end{figure}

In recent years, there has been extensive research on Multimodal Large Language Models (MLLMs), covering foundational model development~\cite{zhu2025internvl3, qwen2.5-VL}, evaluation dataset construction~\cite{yang2025cc, cheng2025comt}, reinforcement learning (RL) applications~\cite{liu2025visual}, and even areas related to AI-Generated Content (AIGC)~\cite{sun2024generative, tian2024visual, huang2025wegen}. At the same time, thanks to the development of diffusion models~\cite{ho2020denoising,rombach2022high,peebles2023scalable,ye2023ip} and unified MLLMs~\cite{huang2025wegen}, the performance of image generation has also been greatly improved. The diffusion process of images involves a certain degree of randomness, so it often requires specific conditions to provide directed control~\cite{zhang2023adding}. However, even with constraints on the generation process, the model may still produce some unpredictable results. Therefore, it is highly necessary to conduct a screening process for the generated images. MLLMs are capable of processing information from different modalities, such as text and image, simultaneously, and providing responses based on a comprehensive understanding. Against this background, this paper focuses on exploring the spatial plausibility reasoning ability of MLLMs for image screening.\\
Spatial plausibility reasoning with MLLMs is hindered by the scarcity of specialized datasets and the suboptimal reasoning abilities of existing models. In response, advancements have been achieved through the development of a novel dataset and an advanced approach. The collected dataset in this paper consists of over 128k samples, providing indispensable knowledge for model training. In contrast to prior studies~\cite{tan2025ominicontrol,zhu2022learning, li2023theme}, our dataset places significant emphasis on the physical space transformations of AI-generated images. These transformations are straightforward to evaluate objectively, avoiding any reliance on subjective artistic criteria. The generated images consist of the foreground medicines, backgrounds, and layout settings. All the medicines are derived from the real world. We randomly select a background image from the background image set and randomly assign a layout (either top-bottom or left-right) for the original medicine image. Then, we use a segmentation model~\cite{BiRefNet} on the original image to obtain the mask of the foreground. Based on the assigned layout region, we determine the position of the foreground target. Finally, we use an image generation model~\cite{flux2024} to render the areas outside the foreground medicine, referencing the selected background image. In the generated images, we observe a significant amount of unintended content due to randomness. As shown in Figure~\ref{Overview} (a), we categorize these issues into four types: appearance deformation, physical shadow, placement layout, and extension rationality. We summarize the ability to identify these four types of issues as spatial plausibility reasoning capability. After collecting the images, we assign the four generated images labels A, B, C, and D, and manually annotate the correct images with multi-answer labels in the training and testing datasets, such as "AC", "BCD", or "N". "N" indicates that there are no matching answers. We test various state-of-the-art closed-source and open-source MLLMs for the identification of problematic images, with the final results presented in Figure~\ref{Overview} (b). From the results, we conclude that directly using existing models yields very poor performance. Based on this, we further propose a two-stage approach to enhance the spatial plausibility reasoning capability of small-sized MLLMs. Specifically, in the first stage, we investigate multiple approaches to acquire CoT data to perform supervised fine-tuning (SFT) on the base model to adapt it to the specific response format. Purely manual annotation produces the highest quality CoT data but incurs high associated costs and a slow pace, making it difficult to scale. Fully automated annotation is extremely low-cost and highly scalable. However, it sacrifices quality, often generating noisy data that contains factual errors or hallucinations. Answer-driven annotation serves as a strategic trade-off. It leverages the final answer from humans and the thought process from existing models to accelerate the process and reduce costs. It balances the efficiency of automation with the reliability of human expertise. In the second stage, we propose  Group Relative Policy Optimization (GRPO)~\cite{shao2024deepseekmath, guo2025deepseek} with Dynamic Proportional Accuracy (DPA) reward, called DPA-GRPO, to stimulate the model's reasoning ability. In DPA-GRPO, the integrated DPA reward provides a significantly denser reward signal to guide the optimization process more effectively. Ultimately, our approach achieves a score of 59.83 on the evaluation dataset with InternVL3-2B, surpassing large-sized open-source and leading closed-source models. We hope our attempts will provide more robust and reliable solutions for multimodal spatial intelligence reasoning. The main contributions of this paper can be summarized as follows:
\begin{itemize}
\item Few existing benchmarks offer multi-answer labels for spatial plausibility visual reasoning. We address this by constructing SPR-128K, a large-scale dataset for spatial plausibility reasoning, built with an image fusion and generation pipeline and annotated with correct answers.
\item We propose a two-stage framework with scalable CoT data acquisition and DPA-GRPO, enabling a 2B model to outperform larger open- and closed-source models by over 20 points on our SPR-128K dataset.
\item Experiments on public datasets reveal the limitations of GRPO in multi-answer reasoning and demonstrate the effectiveness of DPA-GRPO.
\end{itemize}

\section{Related Work}
\subsection{Multimodal Large Language Models}
MLLMs have demonstrated impressive capabilities across various tasks and applications~\cite{zhu2025internvl3, liu2025visual, chen2024far, hong2025glm}. Particularly, an increasing number of studies are exploring the scope and boundary of MLLMs' capabilities. For instance, CC-OCR~\cite{yang2025cc} investigates the performance of MLLMs on end-to-end OCR tasks. Naturalbench~\cite{li2024naturalbench} examines their ability to understand when faced with the same question under different images. CoMT~\cite{cheng2025comt} explores their visualization capabilities during the reasoning process. All-Angle Bench~\cite{yeh2025seeing} studies MLLMs' multi-perspective understanding and corresponding abilities. Inspired by the frequent occurrence of unreasonable content in the field of image generation, we are very curious about whether MLLMs can understand spatial plausibility regarding the generated content. In this work, we construct a pipeline for the image generation dataset and evaluate the performance of various MLLMs based on this dataset.

\subsection{Reinforcement Learning}
Recently, reinforcement learning techniques have been extensively applied to enhance the reasoning capabilities of Large Language Models (LLMs), enabling them to effectively solve complex problems~\cite{shao2024deepseekmath, guo2025deepseek, yu2025dapo, yuan2025vapo, zhang2025gvpo, zheng2025group}. DeepSeek-R1~\cite{guo2025deepseek} is a milestone work in the application of reinforcement learning to the domain of LLMs. It includes the full version of DeepSeek-R1 and DeepSeek-R1-Zero, which are obtained using Group Relative Policy Optimization (GRPO). Based on GRPO, DAPO~\cite{yu2025dapo} introduces several key techniques to make RL shine in the long-CoT RL scenario with Qwen2.5-32B, outperforming the DeepSeek-R1-Zero-Qwen-32B while using only half the training steps in AIME 2024. VAPO~\cite{yuan2025vapo} proposes a novel framework tailored for reasoning models within the value model-based paradigm, pinpointing three key challenges that plague value model-based methods: value model bias, the presence of heterogeneous sequence lengths, and the sparsity of reward signals. As for MLLMs, RL is often applied to specific tasks, such as object detection~\cite{liu2025visual}, training reward models~\cite{wang2025unified}, and enhancing reasoning capabilities~\cite{zhang2025r1}, which often require designing task-specific rewards. In this paper, we apply the reinforcement learning method GRPO to a new task: spatial plausibility reasoning with MLLMs for screening generated images. The proposed DPA-GRPO method, even when applied to small-sized models, outperforms both large-sized open-source and closed-source models.

\section{SPR-128K Dataset}

\begin{figure}[!t]
\centering
\includegraphics[width=0.45\textwidth]{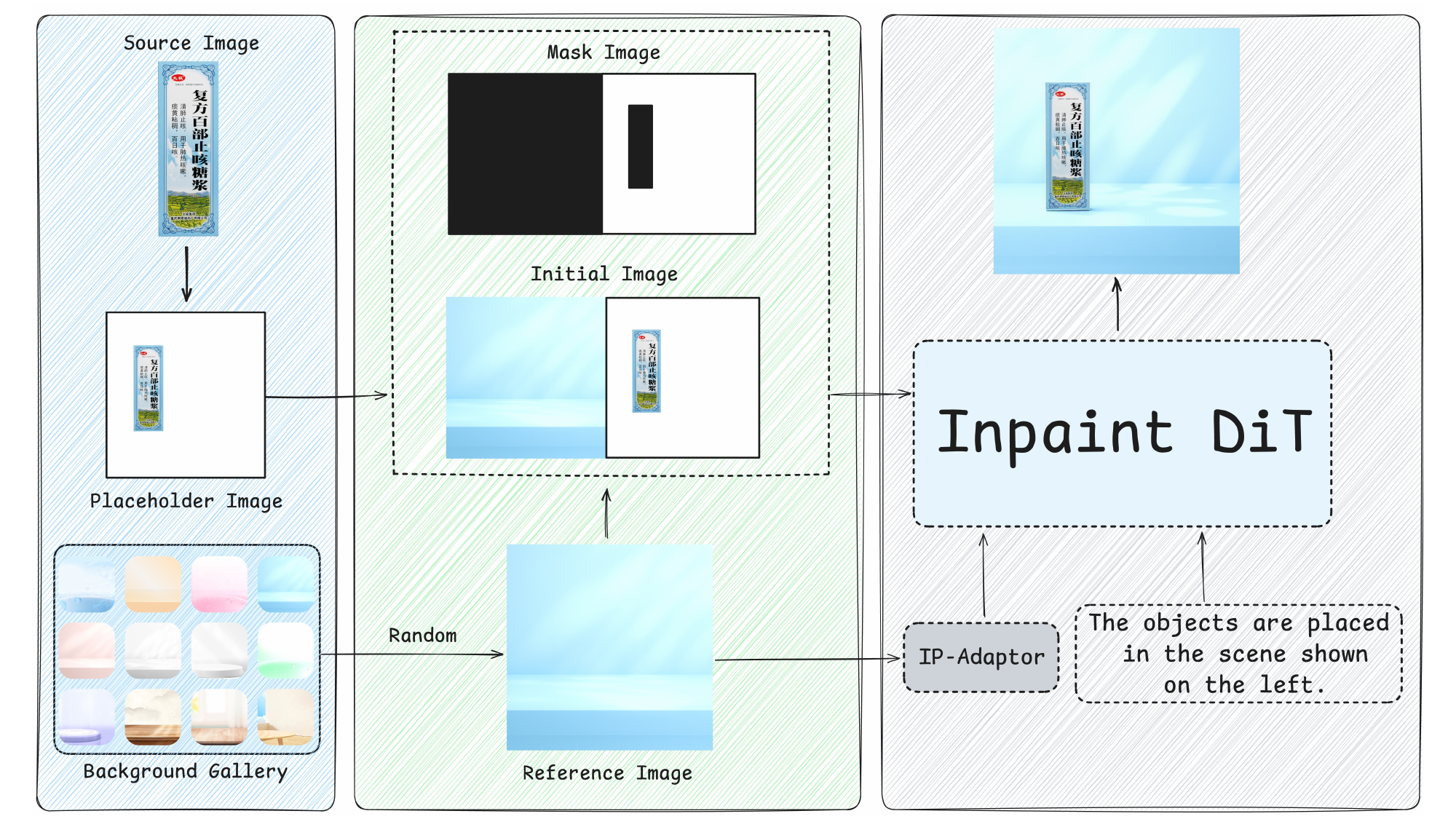}
\caption{Overview of dataset construction pipeline.} \label{Dataset}
\end{figure}

\subsection{Image Fusion and Generation}
In this section, we mainly introduce the data construction pipeline and provide detailed information about the SPR-128K. The construction process can be divided into three stages, as shown in Figure~\ref{Dataset}. The first stage is the data preparation phase. We first obtain source images of various medicines and use an open-source image segmentation model~\cite{BiRefNet} to extract the foreground regions of the medicines. These regions are then placed onto a preset area of a white background image to create the placeholder image. Meanwhile, we randomly select a background image from the background gallery as the reference image. The second stage is the data processing phase. We combine the placeholder image and the reference image to create the initial image. Additionally, we generate a mask image based on the placeholder image. The mask image consists of two parts: the left half is entirely black, while the right half is white, except for the target region. The third stage is the generation phase. Here, we utilize FLUX.1-Fill-dev and FLUX.1-Redux-dev~\cite{flux2024} as Inpaint DiT~\cite{peebles2023scalable} and IP-Adapter~\cite{ye2023ip}, respectively. The reference image is fed into the IP-Adapter to serve as a controller, while the mask image and the initial image are simultaneously input into the Inpaint DiT for repainting. In summary, the process involves redrawing the initial image with the white regions in the mask image by referencing the reference image.\\
Based on the above pipeline, we collect over 640k images grouped into more than 128k samples, covering 56,500 medicine types, 20 backgrounds, and 4 evaluation dimensions. As shown in Figure~\ref{Overview} (a), these dimensions include: (1) appearance deformation, referring to visual inconsistencies with the original medicine; (2) physical shadow, indicating lighting or shadow errors; (3) placement layout, denoting unrealistic spatial arrangements such as floating objects; (4) extension rationality, requiring logically consistent generation without hallucinations that lack any grounding in the original
medicine image. We define the ability to detect these issues as spatial plausibility reasoning capability. 

\begin{table}[!t]
\centering
\caption{An overview of the SPR-128K, detailing their characteristics such as size, label accuracy, CoT data, and supervision type.}
\label{tab:dataset_overview}
\resizebox{\linewidth}{!}{
    \begin{tabular}{ccccc}
    \toprule
    \textbf{Dataset Split} & \textbf{Training} & \textbf{Testing} & \textbf{Pseudo-Label} & \textbf{Exploration} \\
    \midrule
    Size & 1,044 & 468 & 10,724 & 115,809 \\
    Multi-answer Label & \checkmark & \checkmark & $\times$ & $\times$ \\
    CoT Data   & \checkmark & $\times$ & \checkmark & $\times$ \\
    Supervision Type    & Fully + Answer-driven & Fully & Weakly & Unsupervised \\
    \bottomrule
    \end{tabular}
}
\end{table}

\subsection{Dataset Division and Annotation}
\begin{figure}[t]
\centering
\includegraphics[width=0.45\textwidth]{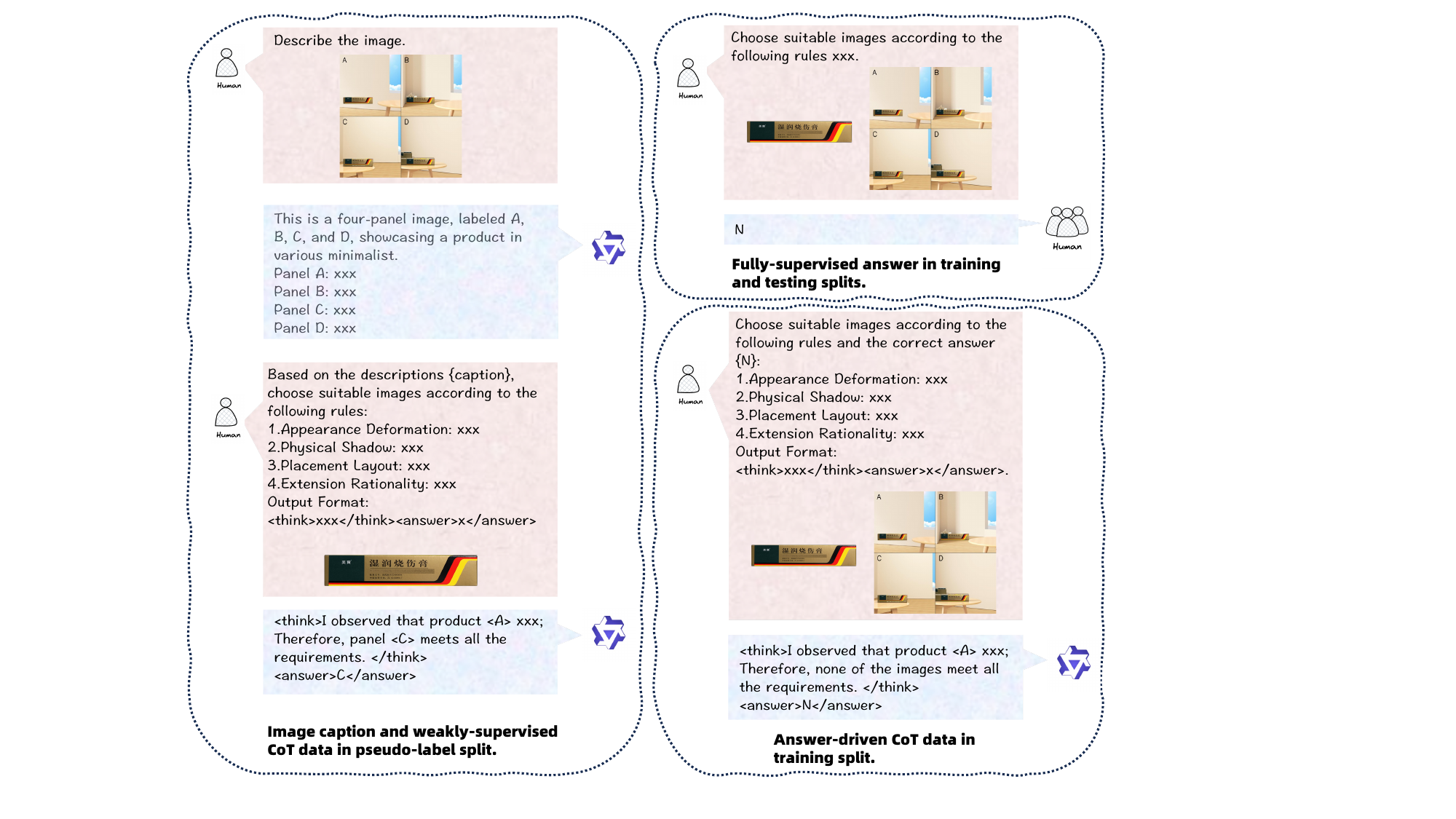}
\caption{Presentation of different annotation paradigms.} \label{data_annotation}
\end{figure}

As shown in Table~\ref{tab:dataset_overview}, our dataset is divided into training, testing, pseudo-label, and exploration splits, totaling 128k unique samples. Each sample includes an original medicine image and four generated images. In the training and testing splits, human reviewers provide accurate manual annotations, as illustrated in the top right part of  Figure~\ref{data_annotation}. Training samples are annotated with multi-answer labels, possibly containing multiple correct options (e.g., “ACD”, “BC”, or “N”). Testing samples include labels for both overall and dimension-specific evaluation, allowing a nuanced assessment of model performance across spatial plausibility dimensions. This annotation process is highly time-consuming, requiring detailed inspection of all four candidate images per sample.\\
To reduce manual effort, we use Qwen-VL-Max~\cite{qwen-vl-max} for automated pseudo-labeling, enhancing the diversity of descriptive tags. As shown in the left part of Figure~\ref{data_annotation}, we adopt a progressive annotation strategy: the model first generates basic image descriptions, leveraging its strong pre-trained captioning ability, then structures reasoning and derives a final answer. Thus, our pseudo-label split contains two forms of annotations: image descriptions and CoT reasoning data.\\
For the final exploration split, we generate four synthetic images from each original image but provide no annotations. This portion of the data is intentionally left unlabeled to serve as a testbed for future exploration of unsupervised methods.

\section{Method}

\subsection{Cold Start: Basic Spatial Understanding and Instruction Following}
\begin{figure*}[!t]
\centering
\includegraphics[width=0.7\textwidth]{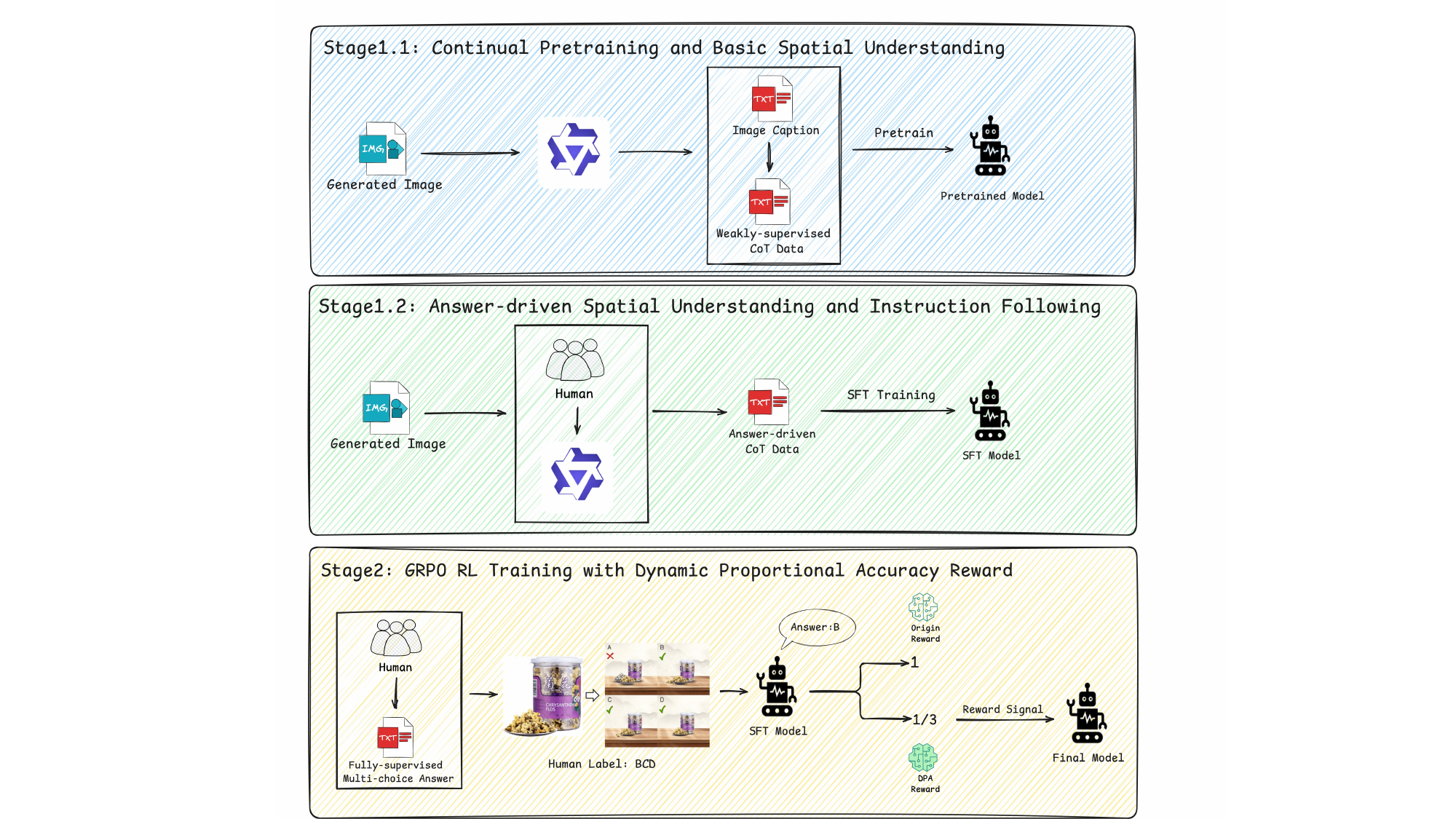}
\caption{Illustration of model training process.} \label{model_training}
\end{figure*}

Due to limited training coverage, MLLMs often violate output formats and exhibit hallucinations, making direct reinforcement learning inefficient. Inspired by DeepSeek-R1-Zero and DeepSeek-R1~\cite{guo2025deepseek}, we introduce a cold-start stage using CoT data before reinforcement learning.
Acquiring CoT data for image screening is costly, as it requires detailed human inspection. To mitigate this, we design two complementary approaches. The first uses the pseudo-label split, where Qwen-VL-Max generates image descriptions and corresponding reasoning steps. Although the weakly supervised CoT achieves only 38.25\% accuracy (Table~\ref{tab:Comparison Results}), it helps the model learn structured responses and basic spatial understanding. The second approach leverages human-labeled training samples to guide Qwen-VL-Max in regenerating 1,044 answer-driven CoT examples, as shown in the bottom right part of Figure~\ref{data_annotation}. These two methods are applied sequentially: continual pretraining on image caption and weak CoT data, followed by instruction-tuned training on answer-driven CoT data, as illustrated in Figure~\ref{model_training} Stage1.1 and 1.2. The objective function of SFT is defined as:
\begin{equation}
\mathcal{L}_{cold\_start}(\theta)=-\sum_{i=1}^T\log p(y_i|x,y_{<i};\theta),
\end{equation}
where $x$ is the original input, $y=\{y_1,y_2,...,y_T\}$ is the distilled output from Qwen-VL-Max, and $\theta$ represents the parameters of the base model. This stage serves to initialize the model's ability to follow a structured CoT reasoning format.

\subsection{DPA-GRPO: Reinforcement Fine-Tuning with DPA Reward}
GRPO removes the critic model and estimates the baseline from group rewards instead~\cite{guo2025deepseek}. Given an input $q$, it samples $G$ responses $\{o_i\}^G_{i=1}$ from the old policy $\pi_{old}$ and evaluates them with predefined reward functions to obtain $\{r_i\}^G_{i=1}$. The baseline is computed as the group mean reward, and normalized advantages are obtained by subtracting the mean and dividing by the standard deviation:
\begin{equation}
A_i=\frac{r_i-mean(\{r_1,r_2,...,r_G\})}{std(\{r_1,r_2,...,r_G\})},
\end{equation}
where $A_i$ quantifies the relative quality of the $i$-th response in comparison to other candidates within the same sampled group. Based on the simple advantage $A_i$, GRPO optimizes the policy model $\pi_{\theta}$ by maximizing the following objective:
\begin{equation}
\begin{split}
\mathcal{J}_{GRPO}(\theta) = \mathbb{E}[q\sim P(Q),\{o_i\}_{i=1}^G\sim\pi_{old}(O|q)] \\
 \frac{1}{G}\sum_{i=1}^G (\min(w_iA_i, clip(w_i,1-\epsilon,1+\epsilon)A_i)  \\
 -\beta\mathbb{D}_{KL}(\pi_\theta||\pi_{ref})), 
\end{split}
\end{equation}

\begin{equation}
\mathbb{D}_{KL}(\pi_{\theta}||\pi_{ref})=\frac{\pi_{ref}(o_i|q)}{\pi_{\theta}(o_i|q)}-log\frac{\pi_{ref}(o_i|q)}{\pi_{\theta}(o_i|q)} -1,
\end{equation}

\begin{equation}
w_i=\frac{\pi_{\theta}(o_i|q)}{\pi_{old}(o_i|q)},
\end{equation}
where $w_i$ is the importance sampling coefficient, $\beta$ controls the deviation from the reference model $\pi_{ref}$, and $\epsilon$ clips extreme importance sampling coefficients for stability.\\
In Deepseek-R1~\cite{guo2025deepseek}, the overall verifiable reward $r$ is formulated as:
\begin{equation}
r=r_{fmt}+r_{acc}, 
\end{equation}
where $r_{fmt}$ and $r_{acc}$ stand for format reward and accuracy reward, respectively. The format reward encourages the model to produce outputs in a fixed structure: $<$think$><$/think$><$answer$><$/answer$>$. We retain this reward by using prompts to guide structured responses. The first-stage CoT learning further ensures stable generation in this format. The accuracy reward measures whether the final answer matches the ground truth. Our DPA reward differs from standard GRPO in its reward design for multi-answer questions. Since a sample may contain multiple correct answers, we introduce a DPA reward, where the score depends on how much of the correct answer is covered:
\begin{equation} 
r_{acc} = 
\begin{cases} 
\frac{|\mathcal{R}|}{|\mathcal{A}|}, & \text{if } \mathcal{R} \subseteq \mathcal{A} \\ 
0, & \text{otherwise} 
\end{cases} 
\end{equation}
where $\mathcal{R}$ is the set of chosen options in the response, $\mathcal{A}$ is the set of ground-truth answers, and $\vert \cdot \vert$ denotes the number of elements in the set. Crucially, the subset condition $\mathcal{R} \subseteq \mathcal{A}$ explicitly handles "mixed" outputs: if $\mathcal{R}$ contains any incorrect choice, it is no longer a subset of $\mathcal{A}$, resulting in a reward of 0. This design grants partial credit only when the response is fully contained within the correct answers, providing a more fine-grained measure of model accuracy.

\section{Experimental Results}

\subsection{Experimental Setting}
We evaluate our model on the proposed spatial plausibility reasoning dataset to assess spatial reasoning and multi-image understanding. The proposed DPA-GRPO demonstrates competitive performance due to the introduction of the dynamic proportional accuracy  reward in the GRPO training scheme.\\
\textbf{Baseline Models.} For the baseline models used for cross-sectional comparison, we select several large-sized open-source models, such as InternVL3.5-30B-A3B~\cite{wang2025internvl3}, InternVL3-78B~\cite{zhu2025internvl3}, Qwen3.5-397B-A17B~\cite{qwen35blog}, Qwen2.5-VL-72B-Instruct~\cite{qwen2.5-VL}, and GLM-4.5V~\cite{hong2025glm}, as well as some closed-source models like GPT5.2~\cite{openai-gpt4o}, GPT4o~\cite{openai-gpt4o}, Gemini3-Pro~\cite{google-gemini3}, Claude Sonnet4.5~\cite{anthropic-claude-sonnet-4-5}, Qwen3.5-Plus~\cite{qwen35blog}, and Qwen-VL-Max~\cite{qwen-vl-max}. For the baseline compact models used for longitudinal comparison, we select InternVL3-1B and InternVL3-2B~\cite{zhu2025internvl3} to validate the effectiveness of our proposed DPA-GRPO.\\
\textbf{Evaluation Metrics.} Due to the high difficulty of the spatial plausibility reasoning task, we differentiate the evaluation metric from the dynamic proportional accuracy reward. Specifically, if the model's response is a subset of the correct options in a multi-answer question, we consider it correct. Our model is designed for a real-world e-commerce task: product image screening. In this scenario, only one image is ultimately displayed on the webpage. Therefore, if the model suggests multiple valid images, we would simply select one from its recommendations. As long as the model provides one correct option, the task is successfully completed, provided that all the output options of the model are correct. In this way, there is no need to identify all the correct options.\\

\begin{table*}[!t]
    \centering
    \caption{Comparison results. We evaluate the image screening performance of both closed-source and open-source MLLMs. We use bold to highlight the top results, and underline to indicate the second-best results. Notably, the overall score includes four dimensions of spatial plausibility ability and the normal type.}
    \label{tab:Comparison Results}
    \renewcommand{\arraystretch}{0.9} 
    \resizebox{\linewidth}{!}{
        \begin{tabular}{c c | c | c c c c}
            \toprule
            \textbf{Models} & 
            \makecell{\textbf{Prompt} \\ \textbf{Type}} & 
            \makecell{\textbf{Overall} \\ \textbf{Score}} &
            \makecell{\textbf{Appearance} \\ \textbf{Deformation}} &
            \makecell{\textbf{Physical} \\ \textbf{Shadow}} &
            \makecell{\textbf{Placement} \\ \textbf{Layout}} &
            \makecell{\textbf{Extension} \\ \textbf{Rationality}} \\
            \midrule
            \multicolumn{7}{c}{\textbf{API-based models}}\\
            \midrule
            \multirow{2}{*}{GPT5.2} & Direct & 50.00 & 46.24 & 51.22 & 53.39 & 49.50\\
                                   & CoT    & 42.95 & 32.09 & 45.12 & 33.90 & 44.55\\
            \multirow{2}{*}{GPT4o} & Direct & 18.59 & 18.66 & 13.41 & 22.88 & 20.59\\
                                   & CoT    & 34.62 & 23.51 & 29.27 & 24.58 & 29.41\\
            \multirow{2}{*}{Gemini3-Pro} & Direct & 44.02 & 33.58 & 39.02 & 37.29 & 33.33\\
                                          & CoT & 37.39 & 27.61 & 34.15 & 26.27 & 25.49\\
            \multirow{2}{*}{ClaudeSonnet4.5} & Direct & 40.17 & 29.59 & 35.80 & 28.45 & 43.14\\
                                          & CoT & 37.82 & 26.59 & 34.15 & 23.73 & 39.22\\
            \multirow{2}{*}{Qwen3.5-Plus} & Direct & 43.59 & 34.33 & 40.24 & 33.90 & 37.25\\
                                         & CoT & 39.74 & 30.60 & 39.02 & 27.97 & 34.31\\
            \multirow{2}{*}{Qwen-VL-Max} & Direct & 32.05 & 21.64 & 26.83 & 25.42 & 15.67\\
                                         & CoT & 38.25 & 29.10 & 37.80 & 25.42 & 28.43\\
            \midrule
            \multicolumn{7}{c}{\textbf{Open-source MLLMs}}\\
            \midrule
            \multirow{2}{*}{GLM-4.5V} & Direct & 34.83 & 23.88 & 29.27 & 27.97 & 26.47\\
                                          & CoT & 32.69 & 20.90 & 30.49 & 22.03 & 23.53\\
            \multirow{2}{*}{Qwen3.5-397B-A17B} & Direct & 43.80 & 35.82 & 37.80 & 33.05 & 39.22\\
                                          & CoT & 39.96 & 31.34 & 35.37 & 31.36 & 32.35\\
            \multirow{2}{*}{Qwen3-VL-30B-A3B-Instruct} & Direct & 33.97 & 23.51 & 25.61 & 25.42 & 26.47\\
                                          & CoT & 36.54 & 27.99 & 36.59 & 22.88 & 30.39\\
            \multirow{2}{*}{Qwen3-VL-8B-Instruct} & Direct & 34.62 & 23.88 & 32.93 & 24.58 & 24.51\\
                                          & CoT & 34.40 & 21.64 & 29.27 & 25.42 & 24.51\\
            \multirow{2}{*}{Qwen2.5-VL-72B-Instruct} & Direct & 39.74 & 29.85 & 40.24 & 32.20 & 30.39\\
                                          & CoT & 35.47 & 25.00 & 35.37 & 25.42 & 25.49\\
            \multirow{2}{*}{InternVL3.5-30B-A3B} & Direct & 32.48 & 20.90 & 26.83 & 22.03 & 26.47\\
                                          & CoT & 27.56 & 18.28 & 21.95 & 22.03 & 25.49\\
            \multirow{2}{*}{InternVL3.5-38B} & Direct & 33.33 & 20.90 & 26.83 & 25.42 & 25.49\\
                                          & CoT & 20.73 & 13.06 & 21.95 & 13.56 & 15.69\\
            \multirow{2}{*}{InternVL3-78B} & Direct & 34.83 & 23.88 & 31.71 & 24.58 & 27.45\\
                                          & CoT & 36.75 & 26.49 & 31.71 & 29.66 & 28.43\\
            \multirow{2}{*}{InternVL3-38B} & Direct & 36.75 & 26.49 & 32.93 & 30.51 & 27.45\\
                                          & CoT & 32.48 & 21.27 & 24.39 & 24.58 & 26.47\\
            \midrule
            DPA-GRPO-1B (Ours) & CoT & \underline{55.56} & \underline{58.96} & \underline{56.10} & \textbf{61.86} & \underline{67.65}\\
            DPA-GRPO-2B (Ours) & CoT & \textbf{59.83} & \textbf{62.31} & \textbf{58.54} & \underline{55.93} & \textbf{69.61}\\
            \bottomrule
        \end{tabular}
    }
\end{table*}

\subsection{Comparison with Prior Arts}
We compare our proposed DPA-GRPO method on small models with several open-source large-sized and closed-source MLLMs. During evaluation, we assess the performance of both direct and CoT format prompts, reporting the overall score as well as the scores for the four evaluation dimensions. The results are shown in Table~\ref{tab:Comparison Results}. In general, all models exhibit suboptimal performance on our dataset. This indicates that these models lack spatial plausibility reasoning capabilities. In this comparative analysis, closed-source models exhibit superior performance relative to open-source alternatives. Notably, GPT5.2 stands out as the most proficient model. For our DPA-GRPO models, DPA-GRPO-2B achieves substantial improvements across most dimensions, particularly in appearance deformation and extension rationality. However, it still struggles with recognizing more advanced physical rules, such as placement layout.

\begin{table}[t]
    \centering
    \caption{Comparison between direct answering and reasoning.}
    \label{tab:abation1}
    \begin{tabular}{c c c}
        \toprule
        \textbf{Models} & \textbf{Prompt Type} & \textbf{Overall Score}\\
        \midrule
        \multirow{2}{*}{InternVL3-1B} & Direct & 30.13\\
                                      & CoT    & 25.85\\
        +SFT & Direct & \textbf{36.75}\\
        +GRPO & CoT & 28.42\\
        \midrule
        \multirow{2}{*}{InternVL3-2B} & Direct & 10.68\\
                                      & CoT    & 23.72\\
        +SFT & Direct & \textbf{45.94}\\
        +GRPO & CoT & 28.42\\
        \bottomrule
    \end{tabular}
\end{table}

\begin{table}[t]
    \centering
    \caption{Effectiveness of different CoT data with InternVL3-1B and InternVL3-2B.}
    \label{tab:abation2}
    \resizebox{\linewidth}{!}{
    \begin{tabular}{c c c c c}
        \toprule
        \textbf{Exp} & 
        \makecell{\textbf{Image Caption} \\ \textbf{Data}} &
        \makecell{\textbf{Weakly-supervised} \\ \textbf{CoT Data}} &
        \makecell{\textbf{Answer-driven} \\ \textbf{CoT Data}} &
        \textbf{Overall Score}\\
        \midrule
        InternVL3-1B &  &  &  & 25.85\\
        Exp1 & $\times$ & $\checkmark$ & $\times$ & 34.19 (+8.34)\\
        Exp2 & $\checkmark$ & $\checkmark$ & $\times$ & 34.62 (+8.77)\\
        Exp3 & $\times$ & $\times$ & $\checkmark$ & 39.96 (+14.11)\\
        Exp4 & $\checkmark$ & $\checkmark$ & $\checkmark$ & \textbf{45.94 (+20.09)}\\
        \midrule
        InternVL3-2B &  &  &  & 23.72\\
        Exp1 & $\times$ & $\checkmark$ & $\times$ & 34.83 (+11.11)\\
        Exp2 & $\checkmark$ & $\checkmark$ & $\times$ & 35.04 (+11.32)\\
        Exp3 & $\times$ & $\times$ & $\checkmark$ & 51.07 (+27.35)\\
        Exp4 & $\checkmark$ & $\checkmark$ & $\checkmark$ & \textbf{53.63 (+29.91)}\\
        \bottomrule
    \end{tabular}}
\end{table}

\subsection{Ablation Studies}
We conduct a series of experiments to evaluate the effectiveness of different modules in our approach based on InternVL3-1B and InternVL3-2B.\\
\textbf{Comparison Between Direct Answering and Reasoning.} Using the training set, we compare the original models, those fine-tuned with direct answers, and those directly optimized with GRPO. As shown in Table~\ref{tab:abation1}, direct answer fine-tuning yields the best scores (36.75 and 45.94), while applying GRPO directly to the base models is ineffective. This suggests that insufficient instruction-following and spatial understanding in the base models hinder the effectiveness of the accuracy-based reward.\\
\textbf{Effectiveness of Different Sources of CoT Data.} As shown in Table~\ref{tab:abation1}, direct fine-tuning with answers provides a certain degree of improvement, while RL-based methods such as GRPO fail when applied to models lacking basic reasoning capabilities. To establish a stronger foundation, we incorporate various forms of CoT data, including image captions, weakly supervised CoT generated by Qwen-VL-Max, and human-annotated answer-driven CoT data, and perform a systematic ablation study, as shown in Table~\ref{tab:abation2}. For InternVL3-2B, integrating weakly supervised CoT  improves the overall score from 23.72 to 34.83, demonstrating that auto-generated reasoning sequences can effectively enhance spatial understanding. The addition of image captions  yields limited gains, increasing the score to 35.04. In contrast, the use of high-quality, answer-driven CoT data  markedly boosts performance to 51.07, highlighting the superior value of guided reasoning supervision. The combination of all data sources  further raises performance to 53.63, indicating a complementary effect between diverse reasoning cues and structured guidance. Comparable results are observed with InternVL3-1B, confirming the generality of these findings.\\
\textbf{Comparison among Different GRPO Paradigms.} From Table~\ref{tab:abation1}, we can infer that directly applying GRPO to the original models is not feasible. Therefore, in this part, we start with the model obtained in Table~\ref{tab:abation2}, which involves SFT with different sources of CoT data. We then progressively incorporate different rewards for comparison: the original binary accuracy reward in GRPO and the DPA reward proposed in this paper, to evaluate the effectiveness of each reward type. To ensure consistency, we utilize the same CoT prompt across all models. From Table~\ref{tab:abation3}, we can conclude that the design of the accuracy-based reward directly impacts the effectiveness of the reinforcement learning method. The performance of the original accuracy reward is inferior to the DPA reward proposed in this paper. This demonstrates that the reward design in reinforcement learning must accurately reflect the intended actions. In the context of the multi-answer question task in this paper, DPA reward aligns most closely with the actual scoring rules. 
For the InternVL3-1B-CoT model, we re-evaluate the performance differences between GRPO and DPA-GRPO. The conclusions are consistent with those from the InternVL3-2B-CoT, confirming that our proposed DPA-GRPO method is also effective on the InternVL3-1B-CoT.\\
\textbf{Generalization to Public Datasets} Given that our dataset focuses on multi-answer spatial plausibility reasoning, we further evaluate DPA-GRPO on public datasets targeting more general multi-answer scenarios, including Mmlu-Multi-Answer~\cite{mmlu-multi} and JEC-QA~\cite{zhong2024agieval}. The Mmlu-Multi-Answer contains 3,362 general multi-answer instances, and the JEC-QA includes 1,999 multi-answer samples. In each dataset, we split it into training and testing sets at a 1:1 ratio and evaluate InternVL3-2B with different fine-tuning methods. As shown in Table~\ref{tab:generalization1}, DPA-GRPO outperforms GRPO by 15.82 points, achieving a score of 61.45 on the Mmlu-Multi-Answer dataset. For this dataset, we conclude that in more general multi-answer scenarios, the DPA reward is more aligned with the scoring criteria, while the sparse reward signal of GRPO is not suitable. Furthermore, the experimental results from the JEC-QA dataset also support this conclusion, confirming the generality of DPA-GRPO in multi-answer scenarios. The smaller margin of improvement for DPA-GRPO over GRPO on the JEC-QA dataset can be attributed to its data composition: JEC-QA only partially contains multi-answer questions, while Mmlu-Multi-Answer is exclusively composed of them.

\begin{table}[t]
    \centering
    \caption{Comparison among different GRPO paradigms.
    }
    \label{tab:abation3}
    \begin{tabular}{c c}
        \toprule
        \textbf{Models} & \textbf{Overall Score}\\
        \midrule
        InternVL3-1B-CoT & 45.94\\
        +GRPO & 53.85 (+7.91)\\
        +DPA-GRPO & \textbf{55.56 (+9.62)}\\
        \midrule
        InternVL3-2B-CoT & 53.63\\
        +GRPO & 58.55 (+4.92)\\
        +DPA-GRPO & \textbf{59.83 (+6.20)}\\
        \bottomrule
    \end{tabular}
    \vspace{-5pt}
\end{table}

\begin{table}[!t]
\centering
\caption{Generalization study on the public datasets.}
\label{tab:generalization1}
\resizebox{\linewidth}{!}{
\begin{tabular}{c c c}
\toprule
\textbf{model} & \textbf{Mmlu-Multi-Answer} & \textbf{JEC-QA}\\
\midrule
InternVL3-2B    & 24.63 & 50.70\\
+SFT            & 34.21 (+9.58) & 59.90 (+9.20)\\
+SFT, +GRPO           & 45.63 (+21.00) & 63.40 (+12.70)\\
+SFT, +DPA-GRPO       & \textbf{61.45 (+36.82)} & \textbf{64.20 (+13.50)}\\
\bottomrule
\end{tabular}}
\end{table}

\begin{figure}[h]
\centering
\includegraphics[width=0.45\textwidth]{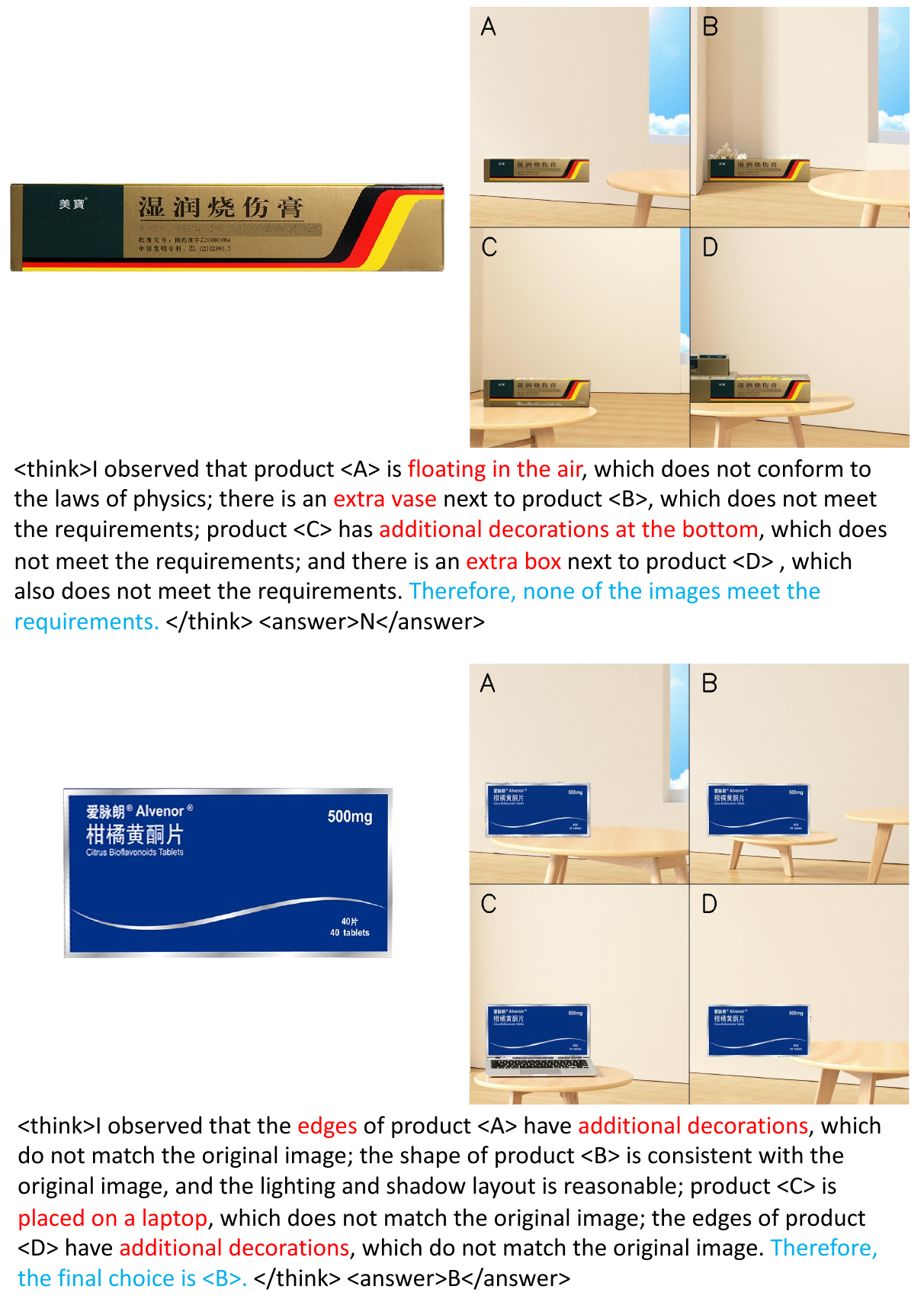}
\caption{Presentation of the reasoning process.} \label{Demonstration2}
\end{figure}

\subsection{Cases Illustration}
In Figure~\ref{Demonstration2}, we illustrate the spatial plausibility reasoning process of the DPA-GRPO-2B model, with key words in the reasoning process highlighted in red and the decisions made by the model marked in blue. As demonstrated by the examples, our DPA-GRPO-2B exhibits robust capabilities for spatial plausibility reasoning.

\section{Conclusion}
In this work, we address the challenges of image screening with MLLMs by proposing a comprehensive solution that includes a novel dataset and an advanced training methodology. Our dataset evaluates spatial plausibility reasoning ability across four critical dimensions, while our method leverages CoT data followed by DPA-GRPO reinforcement learning to significantly enhance performance. Notably, this task is highly challenging, as even leading closed-source MLLMs perform poorly on the test dataset. In contrast, our approach achieves superior results with a much smaller model, demonstrating the effectiveness of combining CoT data with reinforcement learning. Additionally, experiments on more general public multi-answer datasets demonstrate the advantages of DPA-GRPO over the original GRPO. We believe these efforts will provide more robust and reliable solutions for multimodal spatial intelligence reasoning.\\
Despite these achievements, our work still has several limitations. First, our dataset is primarily focused on the medical domain and lacks natural-scene data, which may restrict the generalizability of the model. Future work should therefore incorporate more data from natural scenes to improve its robustness across diverse scenarios. Second, the model shows relatively weak performance in Placement Layout and Physical Shadow. Addressing this issue will require collecting higher-quality CoT data so that the model can better understand spatial relationships.